\documentclass{article}

\usepackage{arxiv}

\usepackage[utf8]{inputenc} 
\usepackage[T1]{fontenc}    
\usepackage{hyperref}       
\usepackage{url}            
\usepackage{booktabs}       
\usepackage{amsfonts}       
\usepackage{nicefrac}       
\usepackage{microtype}      
\usepackage{lipsum}		
\usepackage{graphicx}
\usepackage{natbib}
\usepackage{doi}

\title{Vision Transformers for Mobile Applications: A Short Survey}

\author{Nahid Alam*\\
Cohere for AI Community\\
\and
Steven Kolawole*\\
ML Collective\\
\and
Simardeep Sethi*\\
IIT Delhi
\and
Nishant Bansali\\
Cohere for AI Community\\
\and
Karina Nguyen\\
UC Berkeley\\
}

\renewcommand{\shorttitle}{Vision Transformers for Mobile Applications: A Short Survey}

\hypersetup{
pdftitle={A template for the arxiv style},
pdfsubject={q-bio.NC, q-bio.QM},
pdfauthor={David S.~Hippocampus, Elias D.~Striatum},
pdfkeywords={First keyword, Second keyword, More},
}

\begin{document}
\maketitle
\def\thefootnote{*}\footnotetext{Equal contribution}\def\thefootnote{\arabic{footnote}}

\begin{abstract}
	Vision Transformers (ViTs) have demonstrated state-of-the-art performance on many Computer Vision Tasks. Unfortunately, deploying these large-scale ViTs is resource-consuming and impossible for many mobile devices. While most in the community are building for larger and larger ViTs, we ask a completely opposite question: How small can a ViT be within the tradeoffs of accuracy and inference latency that make it suitable for mobile deployment? We look into a few ViTs specifically designed for mobile applications and observe that they modify the transformer's architecture or are built around the combination of  CNN and transformer. Recent work has also attempted to create sparse ViT networks and proposed alternatives to the attention module. In this paper, we study these architectures, identify the challenges and analyze what really makes a vision transformer suitable for mobile applications. We aim to serve as a baseline for future research direction and hopefully lay the foundation to choose the exemplary vision transformer architecture for your application running on mobile devices.
\end{abstract}


\section{Introduction}
Inspired by the success of transformer \cite{vaswani2017attention} models in Natural Language Processing (NLP) application, A. Dosovitskiy et al. \cite{dosovitskiy2020image} introduced Vision Transformer (ViT) for Computer Vision (CV) applications. Vision transformer models have shown  77.9\% to 81.3\% Top-1 accuracy on ImageNet dataset \cite{touvron2021training},\cite{dosovitskiy2020image},\cite{DBLP:journals/corr/abs-2103-14030} and have been used in many downstream image recognition task such as classification \cite{DBLP:journals/corr/abs-2104-14294},\cite{DBLP:journals/corr/abs-2106-13230}, object detection \cite{DBLP:journals/corr/abs-2005-12872}, \cite{DBLP:journals/corr/abs-2112-01526} and segmentation \cite{DBLP:journals/corr/abs-2105-15203},\cite{DBLP:journals/corr/abs-2112-01527}. Although the Top-1 accuracy scores of these models are impressive, they are generally too big and too slow to deploy on mobile devices \cite{wang2022towards}. These models' massive number of parameters is a burden to compute, storage, memory, and inference latency of a mobile application. The Multi-Head Self-Attention (MHSA) operation in the Transformer module is quadratic in nature.  The MLP module in ViT projects the embedding space by a factor of four, applies non-linearity, and then projects it back to its original shape. As a result, the latency of ViT models, in general, is unrealistic for edge devices such as iPhone, Google Edge \cite{gcloudedge}, NVIDIA Jetson Nano, Intel Edge devices such as \cite{intelmovidiousvpu}, \cite{intelaifpga} and Hexagon DSPs from Qualcomm etc. 

The motivation for this work comes from our experience working with various computer vision applications on mobile devices where CNNs are still prevalent. While the applications in these domains can benefit from the improved performance of a ViT-based backbone, compared to CNN, ViTs have 1.5-4x inference latency \cite{wang2022towards}, \cite{DBLP:journals/corr/abs-2110-02178}. Therefore CNNs are the prevailing architecture of choice in those applications.

Recently, various architectures have been proposed specifically targeting CV tasks in mobile devices that combine transformers with CNNs. In addition, techniques such as quantization, pruning, distillation, etc., are applied to an existing model to fit it in mobile devices \cite{pmlr-v119-kurtz20a}. \cite{Kuznedelev22} demonstrated a training-aware pruning framework to sparsify ViTs that achieves almost SOTA accuracy with 50\% sparsity.

We aim to explore vision transformer architectures optimized for mobile devices. We are specifically interested in their inference efficiency with accuracy/latency tradeoffs. We identify a few common patterns in design choices and challenges in building mobile-optimized ViTs and serve as a guide for edge-optimized vision transformer research.

\section{Related Work}
Han et al. \cite{han2022survey} did a comprehensive survey on Vision Transformers. Still, their work focused on the traditional ViT architectures, not on the ones specifically suitable for mobile applications. Wang et al. \cite{wang2022towards} looked into the state of Vision Transformer for mobile devices from the inference perspective. Their work is designed to check if traditional ViTs are suitable for mobile or not and have not looked into any mobile-suitable ViTs other than T2T-ViT \cite{Yuan_2021_ICCV}. In our work, we look into 8 different Vision Transformers designed for mobile applications, pruning frameworks for sparsifying ViTs, and alternatives to attention networks that can further enhance inference efficiency for Vision Transformers. 

\section{Vision Transformers for Mobile Applications}
\label{sec:headings}

Mobile suitable vision transformers can be broadly classified into two categories. First, changing the model post-training - techniques such as quantization, pruning \cite{han2015learning}, \cite{han2015deep} and a mix of model sparsification techniques \cite{pmlr-v119-kurtz20a} are applied. Second - model architecture changes. This can be a mix of transformer+CNN model, MLP-based models \cite{touvron_resmlp}, or sparse models \cite{fedus2022}. This work focuses on the architectural changes that make vision transformers suitable for mobile applications.

\subsection{Exploring the Architectures}
\paragraph{Receptive Field Increase} A common challenge in CV tasks is to process higher-resolution images for better accuracy without additional compute costs. Both Efficient-ViT \cite{cai2022efficientvit} and EdgeNeXt \cite{maaz2022edgenext} propose design changes to tackle this issue. Efficient-ViT consists of MobileNetV2 \cite{sandler2018mobilenetv2} Depth-Wise Convolution (DWConv)  layers followed by a linear attention module that replaces Softmax attention. It is followed by a Feed Forward Network (FFN) with deformable CNN. Similarly, EdgeNeXt proposes a Split Depth-wise Transpose Attention (SDTA) encoder instead of the vanilla MHSA module. The SDTA encoder splits the input tensors into multiple channel groups. It then utilizes DWConv and self-attention across channel dimensions to effectively increase the receptive field the model operates on.

\paragraph{Pooling Layers}A set of hybrid CNN+Transformer architectures uses pooling layers after the attention module to decrease the inference latency. NextViT \cite{li2022next} consists of alternate Convolutional and Transformer blocks. The Convolutional block uses Multi-head Convolution Attention (MHCA) formed of a grouped 3X3 Convolution operation and an MLP. The Transformer block uses MHSA consisting of a pooling layer for the Key and Value vectors. Moreover, channel reduction is applied to improve inference speed. PoolFormer \cite{yu2022metaformer} suggests an alternative token mixture module -  MetaFormer, where the token mixture is modified with a pooling layer, resulting in a similar accuracy  with fewer parameters compared  to ViT.

\paragraph{Principles of CNN}LeViT \cite{graham2021levit} brings the principles of CNN network to transformers - specifically activation maps through decreasing resolution. It passes the input through a 3x3 convolution and then does a shrinkage of the attention module. Instead of positional embeddings, LeViT encodes the location information through an attention bias term to the attention map. The training process of LeViT is similar to DeiT, which uses a distillation token head after the last stage. MobileViT \cite{DBLP:journals/corr/abs-2110-02178} introduces the MobileViT module that is responsible for capturing the interaction of local and global features of the image. The local representation is captured using NxN convolutions with 1x1 point-wise convolution whose output feature map serves as an input for the global feature capturing module.

\begin{figure}
	\centering
    {\includegraphics[width=\linewidth]{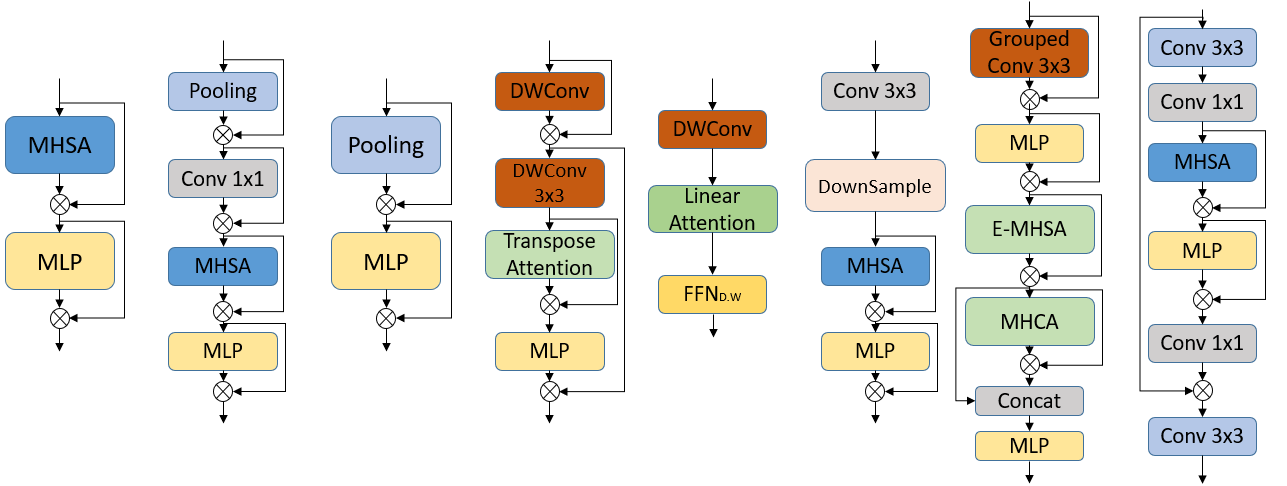}}
	\caption{Abstract Vision Transformer Architectures From Left to Right: Vanilla Transformer;
Efficient-Former; PoolFormer; EdgeNext; EfficientViT; LeViT; NextViT; MobileViT}
	\label{fig:architecture}
\end{figure}

\paragraph{Consistent Tensor Dimension}EfficientFormer \cite{li2022efficientformer} brings a consistent feature dimension to the token mixture. The overall network is divided per the tensor dimension; First, based on 4D tensors. A Convolution layer termed MB4D is applied on 4-Dimensional tensors and, second, based on 3D tensors  consisting of MHSA as a token mixer for the transformer module. Embedding layers are used between stages to project the token length to a lower dimension, alleviating the dimension mismatch and thereby improving the inference speed. Figure \ref{fig:architecture} shows a high-level view of these architectures described so far.

\paragraph{Parallel Layers} We found a category of models that execute specific layers in parallel to improve time complexity. The two parallel branches of Mobile-Former \cite{chen2022mobile} extract local and global representations, respectively and concatenate through a bridge architecture. MixFormer's \cite{chen2022_mixformer} two parallel branches communicate with each other through channel interactions and spatial interactions, providing complementary clues for better representation learning in both branches.

\paragraph{Token Reduction.} Memory and inference latency in a transformer strongly correlate to the number of tokens. But many of these tokens are redundant for the task at hand. Recent work has demonstrated token reduction by pruning tokens \cite{meng_adavit_2022}, \cite{fayyaz_adaptivetoken_2022}, combining tokens \cite{kong_spvit_2021}, \cite{ryoo2021tokenlearner}. All of these methods require training of the model, while ToMe \cite{bolya_tome_2022} proposes a technique for merging tokens based on similarity without any further model training. 
    
\paragraph{Integer Quantization} Quantization is used to decrease inference latency at edge devices. However, an integer-only arithmetic operation on the vision transformer is an open challenge. I-ViT \cite{li2022vit} addresses this gap by applying a dyadic arithmetic pipeline to use integer-only arithmetic for linear operations. For non-linear operations, I-ViT introduces 3 novel operations - Shiftmax, ShiftGelu, and I-LayerNorm. Shiftmax serves as an alternative to the softmax function. It uses the scaling factor of quantization obtained from the MatMul operation and the quantized integer weight to get an integer-only value. Similarly, ShiftGelu is an alternative to the Gelu activation function and I-LayerNorm for an integer-only layer normalization operation that improves top-1 accuracy and latency.

\subsection{Sparse Vision Transformers}
One approach to building an optimized Vision Transformer is to build a Sparse Neural Network (SNN) version of the dense model. SNN typically refers to \cite{liu_2023}, \cite{gale_2019} weight sparsity, i.e. setting weights to zero. This can be helpful when the model is over-parameterized for a given task. SNN can also be based on activation sparsity which measures the number of zeros after non-linearity, i.e, activation function \cite{rhu_2018}.

Sparsity can be structured - where a block of weight matrix can either be dense or sparse \cite{gale_2019}. In unstructured sparsity, there is no pattern to follow \cite{kuzmin_2019}, \cite{liu_2023}. Unstructured sparsity removes the least significant weights regardless of where they are - leading to irregular sparse patterns. Han et el. \cite{han2015deep} proposed the popular method of weight pruning at Neural Networks - training, pruning, and retraining. This approach generally requires multiple iterations of retraining to recover the lost accuracy. This is specifically challenging for Vision Transformers since they are already hard to train and stabilize. 

While CNNs allow for greater than 80\% unstructured sparsity \cite{singh_2020}, and up to 40\% structured sparsity \cite{liu_groupfisher_2021}, before significant accuracy drops, ViT sparsity is harder to achieve. Kuznedelev et al. \cite{kuznedelev_ovit_2022} showed that ViT models lose significant amounts of accuracy in each pruning step. And recovering that lost accuracy is harder for ViTs than CNN. 
To alleviate this issue, Frankle et al. in Lottery Ticket Hypothesis [LTH] \cite{frankle_lth_2018} showed that it is possible to find a Sparse subnetwork (i.e. “winning tickets”) inside a  Dense network that we can train in isolation for full accuracy. Finding that sparse subnetwork, though, is an NP-hard problem, therefore, requires an iterative pruning process to get to that useful subnetwork. Chen et al. \cite{chen_chasingsp_2021} demonstrated the use of an end-to-end dynamic sparse network to address the challenges of LTH. The authors used ImageNet with DeiT-Tiny/Small/Base. They demonstrated 50\% model sparsity with DeiT-Small produces, saving 49.32\% FLOPs and 4.40\% running time while attaining a surprising improvement
of 0.28\% accuracy. Kuznedelev et al. \cite{kuznedelev_ovit_2022} proposed the oViT framework that iteratively prunes the model while finetuning. With oViT, EfficientFormer-L1 with 50\% sparsity achieves 78\% Top-1 accuracy vs  the dense counterpart of 78.9\% accuracy.

\subsection{Alternatives to Attention}

Recent work on finding alternatives to the attention module shows promising results. Zhai et al. \cite{zhai_attentionfree_2021} proposed an attention-free transformer architecture. The authors suggest replacing attention with gating and Softmax or gating and explicit convolution. Michael et al. \cite{poli_hyena_2023} look to find a subquadratic operator that can match the quality of the attention network at scale. They showed that their attention-free network reaches transformer quality with 20\% less compute for language modeling tasks. Their proposed subquadratic Hyena operators are twice as fast as highly optimized attention at sequence length 8K and 100× faster at sequence length 64K, tested on WIKITEXT103 and THE PILE dataset. Using these subquadratic operators as a drop-in replacement for attention modules can be a direction for future research.

\subsection{Comparing Results}

\begin{figure}
	\centering
    {\includegraphics[width=\linewidth]{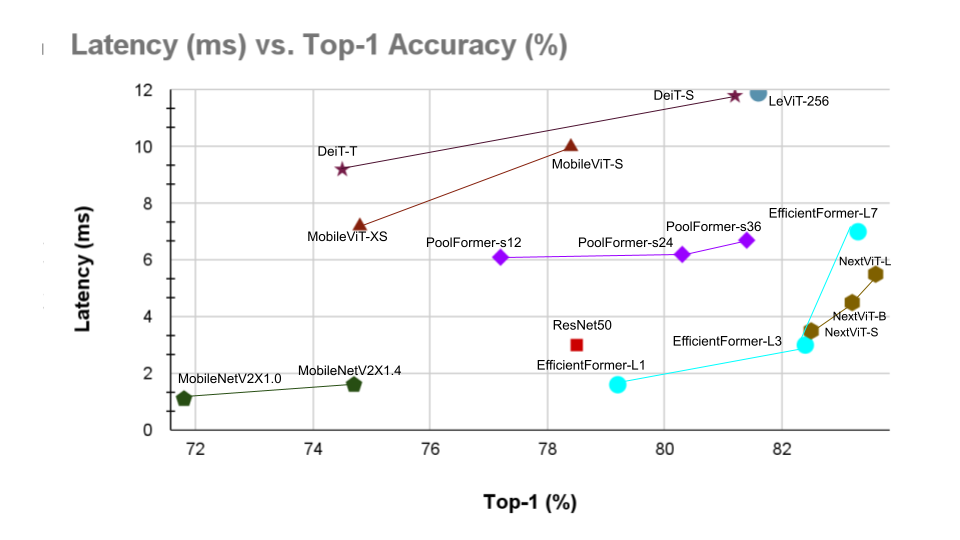}}
	\caption{Top-1 Accuracy vs. Latency for Image Classification on ImageNet-1K using CoreML on iPhone12}
	\label{fig:accuracylatency}
\end{figure}

Figure \ref{fig:accuracylatency} shows Top-1 accuracy vs. latency on several Vision Transformer models that might be suitable for edge devices. These models are  trained on ImageNet-1K for 300 epochs with AdamW optimizer on an image resolution of 224x224. Latency is measured using the CoreML framework on an iPhone 12. Table \ref{tab:example} lists a few state-of-the-art models based on CNN, Hybrid, and pure transformer-based architectures and their number of parameters.

\begin{table}
  \centering
  \begin{tabular}{@{}lc@{}}
    \toprule
    Model & Parameters(M)\\
    \midrule
    ResNet50 & 25.5 \\
MobileNetv2 & 6.1  \\
DeiT-T &  5.9 \\
DeiT-S & 22.5  \\
EdgeNeXt & 5.6\\
MobileViT-XS & 2.3\\
LeViT-256 & 18.9\\
EfficientFormer-L1 & 12.3\\
EfficientFormer-L3 & 31.3\\
EfficientFormer-L7 & 82.1\\
MobileFormer-214 & 9.4\\
PoolFormer-s12 & 12\\
PoolFormer-s24 & 21\\
PoolFormer-s36 & 31\\
EfficientViT & 10.9\\
NextViT-S & 31.7\\
NextViT-B & 44.8\\
NextViT-L & 57.8\\
    \bottomrule
  \end{tabular}
  \caption{Number of model parameters on ImageNet classification}
  \label{tab:example}
\end{table}

These results suggest that we have a few interesting choices for mobile vision transformers within the accuracy-latency tradeoffs. For example, EfficientFormer-L3 has an 82.4\% accuracy with a considerably low latency of 3ms. In comparison, NextViT-S has an 82.5\% accuracy on the classification task, with a 3.5ms latency on CoreML. In our observation, a sub-20M parameter model is a mixed CNN-transformer architecture, mainly because the sub-quadratic alternatives to attention operators are still not prevalent in ViTs \cite{poli_hyena_2023}. 

\section{Conclusion and Future Work}

We aim to serve as a bird's-eye view of the current state of ViTs for mobile applications. We surveyed ViTs that are specifically designed for mobile applications. We identified a few different models with around 10M parameters and reasonable latency-accuracy trade-offs. The majority of these models reduce the time complexity of the MHSA module by modifying the attention operation or completely removing it. ViT one-shot pruning is hard, and the accuracy recovery process is challenging \cite{kuznedelev_ovit_2022} - leading to the need for further research. Token reduction techniques, specifically the ones without requiring further model training, might serve the purpose of the mobile vision transformer community. Moreover, we think that finding attention-free subquadratic operators that act as a drop-in replacements for attention modules in transformers \cite{poli_hyena_2023} can be a research direction worth exploring for mobile vision transformers.




\newpage
\bibliographystyle{unsrtnat}
\bibliography{references}  






\end{document}